\documentclass{article}

\usepackage{arxiv}
\usepackage{tikz}
\usetikzlibrary{shapes.geometric, arrows.meta, positioning, shadows}

\tikzstyle{block} = [rectangle, rounded corners, minimum width=3.2cm, minimum height=1.2cm, text centered, draw=black, fill=gray!10]
\tikzstyle{arrow} = [thick, -{Latex[width=2mm,length=2mm]}]

\usepackage[utf8]{inputenc}     % allow utf-8 input
\usepackage[T1]{fontenc}        % use 8-bit T1 fonts
\usepackage{url}                % simple URL typesetting
\usepackage{booktabs}           % professional-quality tables
\usepackage{amsfonts}           % blackboard math symbols
\usepackage{microtype}          % microtypography
\usepackage{graphicx}           % figures
\usepackage{amsmath}            % for \text in math mode
\usepackage{natbib}             % bibliography
\usepackage{enumitem}
\usepackage{pgfplots}
\pgfplotsset{compat=1.18}
\usepackage{fancyhdr}
\usepackage{placeins}
\usepackage[justification=centering, singlelinecheck=false]{caption}
\usepackage[hidelinks,colorlinks=true,citecolor=blue,linkcolor=blue,urlcolor=blue]{hyperref}

\renewcommand{\arraystretch}{1.2}
\allowdisplaybreaks[2]

\title{One Supervisor, Many Modalities: Adaptive Tool Orchestration for Autonomous Queries}

\author{%
  \begin{tabular}[t]{c@{\hskip 2em}c}
    \href{https://orcid.org/0009-0002-9443-5621}{\includegraphics[scale=0.06]{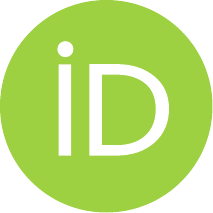}\hspace{1mm}Mayank Saini}\textsuperscript{\dag} &
    \href{https://orcid.org/0000-0001-8064-2035}{\includegraphics[scale=0.06]{orcid.pdf}\hspace{1mm}Arit Kumar Bishwas}\textsuperscript{\dag}\thanks{$\ast$Corresponding author. \textsuperscript{\dag}These authors contributed equally.} \\
    PwC US & PwC US \\
    \texttt{mayank.s.saini@pwc.com} & \texttt{arit.kumar.bishwas@pwc.com}
  \end{tabular}
}

\date{}

\hypersetup{
  pdftitle={One Supervisor, Many Modalities: Adaptive Tool Orchestration for Autonomous Queries},
  pdfsubject={Agentic AI, Multimodal AI, Centralized Orchestration, Cost-Aware Inference},
  pdfauthor={Mayank Saini, Arit Kumar Bishwas},
  pdfkeywords={Agentic AI, AI Agents, Multimodal AI, Multimodal Reasoning, AI Orchestration, Tool Orchestration, Tool-Augmented LLMs, Cost-Aware Inference, LLM Routing, Multi-Agent Systems, Autonomous Query Processing}
}

\begin{document}

\maketitle

\begin{abstract}
We present an agentic AI framework for autonomous multimodal query processing that coordinates specialized tools across text, image, audio, video, and document modalities. A central Supervisor dynamically decomposes user queries, delegates subtasks to modality-appropriate tools (e.g., object detection, OCR, speech transcription), and synthesizes results through adaptive routing strategies rather than predetermined decision trees. For text-only queries, the framework uses learned routing via RouteLLM, while non-text paths use SLM-assisted modality decomposition. Evaluated on 2,847 queries across 15 task categories, our framework achieves 72\% reduction in time-to-accurate-answer, 85\% reduction in conversational rework, and 67\% cost reduction compared to the matched hierarchical baseline while maintaining accuracy parity. These results demonstrate that intelligent centralized orchestration fundamentally improves multimodal AI deployment economics.
\end{abstract}


\section{Introduction}

Modern AI deployment confronts a critical challenge in reconciling conflicting requirements that users demand from production systems: the ability to autonomously process any query type ranging from simple text instructions to complex multimodal requests involving images, audio, video, and structured documents, while simultaneously maintaining operational cost-efficiency at scale and delivering real-time responsiveness suitable for interactive applications. Current solutions prove inadequate along multiple dimensions. Monolithic large language model deployments that route all queries to a single powerful model such as GPT-4 \cite{openai2023gpt4} or Gemini Ultra \cite{gemini2023} incur prohibitive operational costs when applied uniformly across heterogeneous workloads, particularly given that the vast majority of real-world queries do not require the full reasoning capacity of frontier models \cite{ding2024hybridllm,chen2024frugalgpt}. Conversely, hierarchical routing systems that attempt to direct queries through predetermined decision trees based on explicit classification rules exhibit catastrophic brittleness when queries deviate from anticipated patterns---requiring complete pipeline restarts that waste computational resources already invested in partial execution, creating unacceptable latencies for time-sensitive applications, and producing frustrating user experiences that undermine system trust and adoption.

\begin{figure*}[t]
\centering
\includegraphics[width=\textwidth]{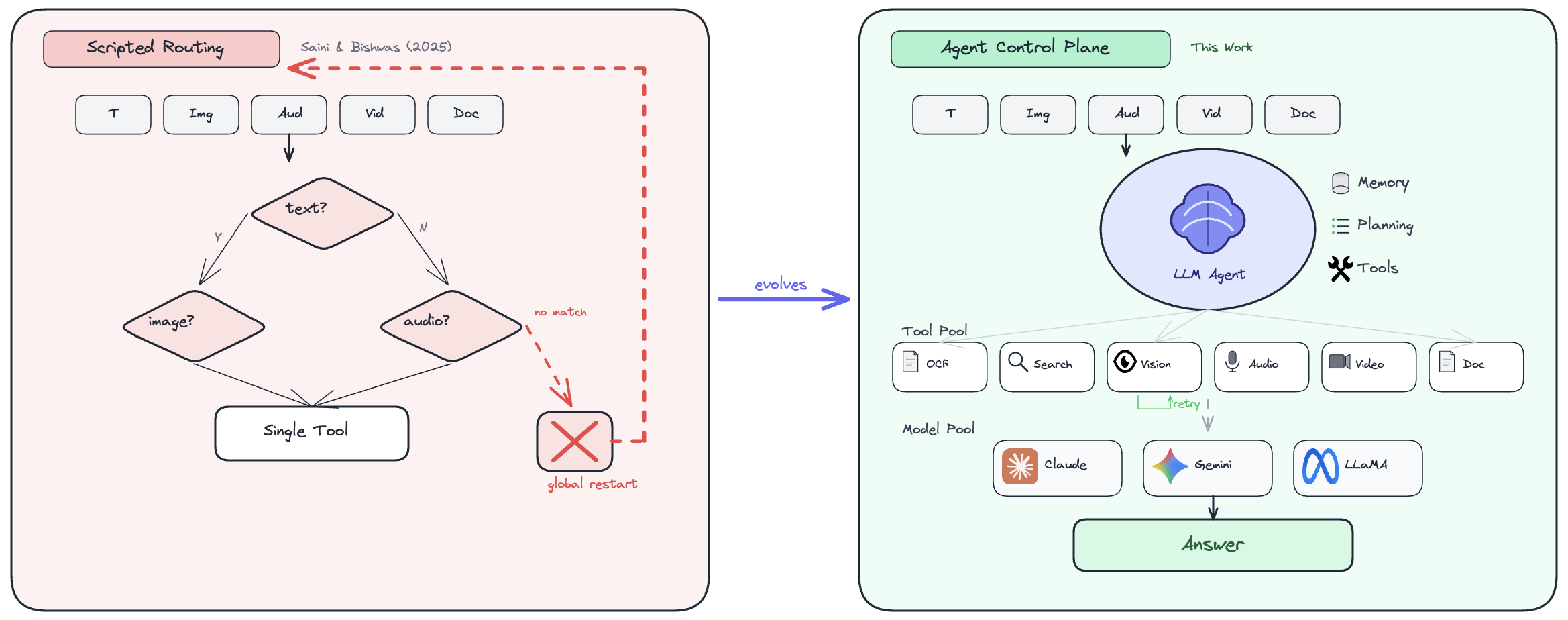}
\caption{Architectural paradigm comparison showing the transition from scripted routing with brittle global restarts to Supervisor-driven adaptive orchestration with tool and model pools.}
\label{fig:paradigm}
\end{figure*}

Our earlier work on resource-efficient multimodal intelligence \cite{saini2025resourceefficient} established the fundamental viability of coordinated processing across diverse modalities through conditional orchestration strategies that route queries based on detected input characteristics. However, that mechanistic approach relied on manually specified routing logic that required explicit enumeration of all anticipated query types and their associated processing paths, creating substantial maintenance burdens as new capabilities were added and producing brittle failure modes when real-world queries fell outside the anticipated design space, as illustrated in Figure~\ref{fig:paradigm}. When users submitted queries that did not match any predetermined pattern---whether due to novel phrasing, unexpected modality combinations, or edge cases not considered during system design---the conditional orchestration framework lacked any mechanism for graceful degradation or adaptive response, instead failing entirely and forcing users to manually reformulate their requests. This fundamental limitation arises directly from the architectural decision to embed all routing intelligence within predetermined decision trees rather than enabling the system to reason about and adapt to query characteristics autonomously based on learned patterns and contextual understanding.

We introduce a centralized orchestration framework that reimagines multimodal query processing through intelligent coordination of specialized tools. The framework employs a central Supervisor that reads tool specifications expressed as typed interfaces with preconditions, postconditions, and latency priors, makes contextual routing decisions based on query characteristics and historical memory state, and dynamically decomposes tasks for delegation to appropriate tools. This approach addresses the limitations of both monolithic systems (which process all queries uniformly at high cost) and hierarchical routing (which relies on predetermined decision trees). For perceptual workloads, the Supervisor routes to domain-optimized tools (e.g., detection, embedding, OCR, transcription) and then contextualizes their outputs for downstream synthesis.

We validate our framework through comprehensive evaluation across 2,847 queries spanning 15 distinct task categories drawn from standard benchmarks. Compared to a matched hierarchical baseline, our approach achieves 72\% median reduction in time-to-accurate-answer with interquartile range of 65--77\%, 85\% reduction in conversational rework requiring user clarification or correction, 67\% reduction in expensive model invocations, and 20\% improvement in concurrent throughput (54 vs 45 q/s) under realistic load conditions, all while maintaining accuracy parity within statistical variance. Perceptual tasks benefit from routing to specialized models (e.g., object detection in 180 ms/frame) instead of end-to-end LLM vision approaches (e.g., 2.4 s/frame), reducing latency and cost while preserving accuracy. These results establish that intelligent orchestration through composable supervision can fundamentally reshape the economics and scalability of multimodal AI deployment without sacrificing response quality or system reliability.

\section{Related Work}
\label{sec:related}

Our work intersects with four active research areas: agentic workflows for tool coordination, multi-agent orchestration frameworks, cost-aware model routing strategies, and multimodal intelligence systems. Recent advances demonstrate large language models' growing capability to interact with external tools and environments through learned action selection. ReAct \cite{yao2022react} establishes a foundation by interleaving reasoning traces with environment actions to reduce hallucination in complex decision-making tasks. Toolformer \cite{schick2023toolformer} extends this capability by learning when to invoke external APIs through self-supervised training on annotated demonstrations. MRKL \cite{karpas2022mrkl} provides theoretical foundations for composing neural and symbolic modules within unified architectures, demonstrating that hybrid systems can leverage complementary strengths of learned and engineered components. HuggingGPT \cite{shen2023hugginggpt} positions large language models as meta-controllers that decompose user requests into subtasks, dispatch each subtask to appropriate specialist models from repositories like Hugging Face, and synthesize partial results into coherent final responses. Concurrent model-centric multimodal systems, including GPT-4, Gemini, and open vision-language models such as LLaVA, substantially improve end-to-end capability but do not directly address orchestration-level cost and repair policies \cite{openai2023gpt4,gemini2023,liu2023llava}. While these approaches demonstrate effective tool utilization, they maintain fixed orchestration architectures where the coordination logic remains embedded within the system. Our framework builds on these concepts by implementing a flexible centralized orchestrator that coordinates specialized tools through context-conditioned routing strategies rather than predetermined decision trees.

Multi-agent orchestration frameworks provide infrastructure for coordinating multiple specialized components within complex workflows. AutoGen \cite{wu2023autogen} introduces conversational patterns where agents negotiate through structured dialogue to resolve ambiguous requests and dynamically adjust execution strategies based on intermediate results. LangGraph \cite{langgraphSite} extends this coordination through graph-based workflow specifications that support branching, conditional execution, and cyclic dependencies. MegaAgent \cite{wang2024megaagent} demonstrates large-scale coordination across hundreds of specialized agents without requiring predefined standard operating procedures, instead relying on learned negotiation protocols. Recent surveys \cite{talebirad2024aiagent,xu2024multimodalagents} highlight rapid progress in agent reasoning, planning capabilities, and the emergence of increasingly sophisticated coordination protocols. ConAgents \cite{zhuang2024conagents} introduces cooperative tool learning where agents jointly discover optimal tool combinations through multi-agent reinforcement learning. However, existing frameworks maintain hierarchical architectures where a central controller makes coordination decisions. Our centralized orchestration approach implements an intelligent orchestrator that dynamically decomposes tasks, delegates to specialized tools, and synthesizes results---enabling parallel processing, adaptive tool selection, and local failure recovery through context-conditioned routing rather than predetermined workflows.

Cost-aware routing and model selection strategies address the economic challenges of deploying large language models at scale. FrugalGPT \cite{chen2024frugalgpt} establishes a framework for adaptive model cascades that route queries through progressively more capable models until quality thresholds are satisfied, achieving up to 98\% cost reduction on certain workloads while maintaining acceptable accuracy. RouteLLM \cite{ong2024routellm} trains learned classifiers on human preference data to predict when queries benefit from powerful models versus when smaller alternatives suffice, demonstrating over 50\% reduction in expensive model invocations on benchmarks like GSM8K and MT-Bench \cite{cobbe2021gsm8k,zheng2023mtbench}. Hybrid LLM \cite{ding2024hybridllm} introduces confidence-based verification where smaller models attempt initial responses and escalate uncertain cases to stronger models, reducing premium usage by approximately 40\% with minimal latency overhead. Cross-attention routing \cite{pulishetty2025crossattention} explores sophisticated attention mechanisms for dynamic model selection based on query-specific features. While these approaches demonstrate substantial cost savings, they focus exclusively on text-only scenarios and optimize narrow metrics like per-query expense rather than holistic user experience. Our framework extends cost-aware routing to multimodal contexts where perceptual decoding imposes baseline computational costs regardless of query complexity. Rather than optimizing cost in isolation, we optimize time-to-accurate-answer---a composite metric that accounts for both response latency and rework probability---across all input modalities. The approach naturally reduces costs by minimizing expensive model invocations while prioritizing the user experience dimensions that most strongly affect system utility and adoption.

Our earlier work on resource-efficient multimodal intelligence \cite{saini2025resourceefficient} demonstrated the viability of coordinated processing across diverse modalities through conditional orchestration strategies. That framework routes queries based on detected input characteristics using manually specified decision trees that enumerate anticipated patterns and their associated processing paths. Conditional systems require explicit specification of all routing paths, creating substantial maintenance burdens as capabilities expand and producing brittle failure modes when real-world queries fall outside the anticipated design space. When users submit queries with novel phrasing, unexpected modality combinations, or edge cases not considered during system design, conditional orchestration lacks mechanisms for graceful degradation or adaptive response---instead failing entirely and forcing users to manually reformulate requests. Our centralized orchestration approach eliminates this brittleness by treating coordination as a learned capability. The central orchestrator reads tool specifications expressed as typed interfaces and makes contextual routing decisions based on query characteristics, historical memory state, and real-time feedback signals. This architectural shift enables handling novel patterns without manual intervention, composing optimal tool combinations dynamically rather than following fixed pathways, and recovering from failures through local repair mechanisms that address specific tool failures without affecting other pipeline components or requiring complete restarts.

\section{Centralized Orchestration Architecture}
\label{sec:architecture}

\begin{figure*}[t]
\centering
\includegraphics[width=\textwidth,trim=60 20 80 50,clip]{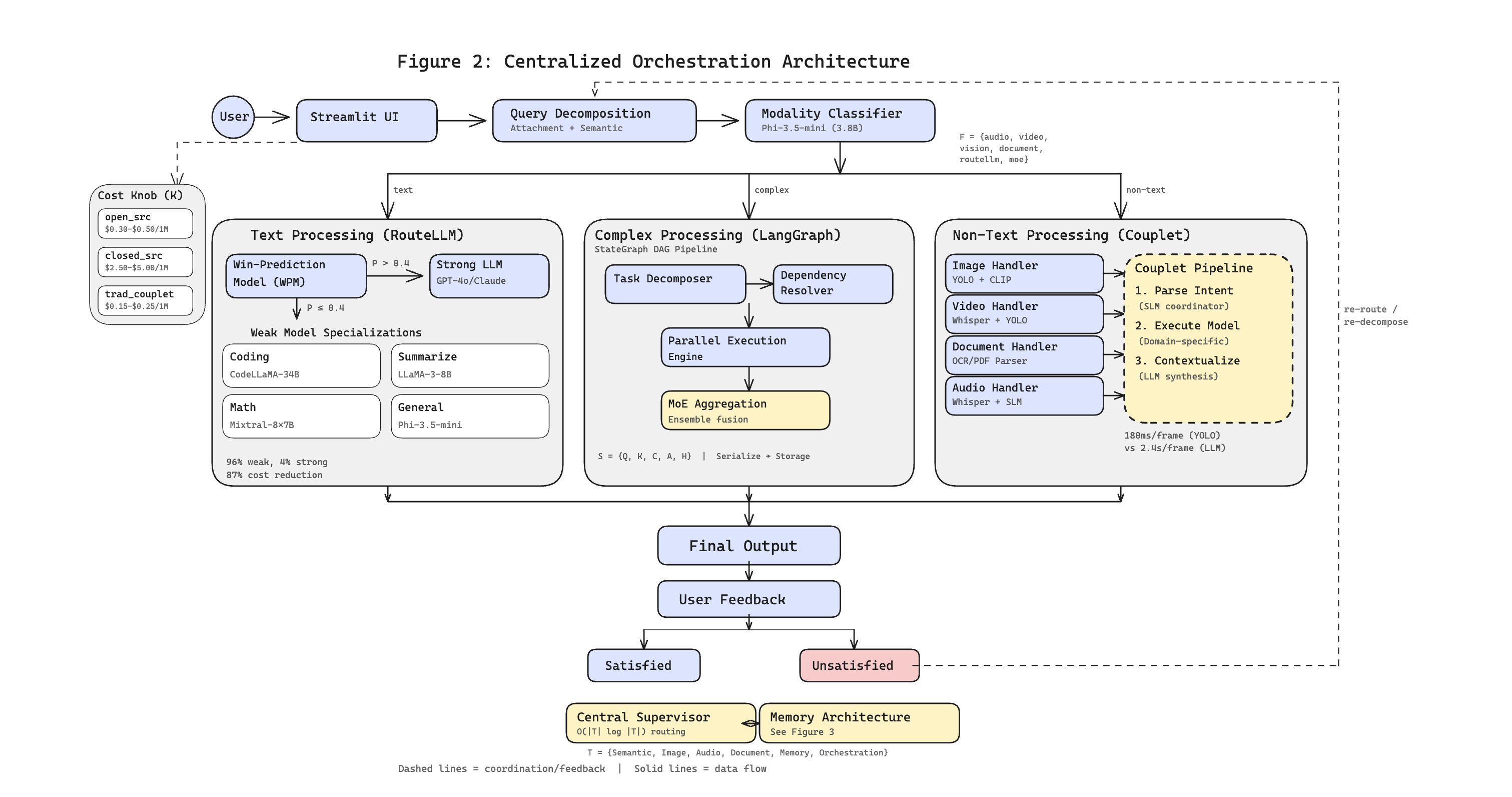}
\caption{Centralized orchestration architecture coordinating specialized tools across modalities through dynamic task decomposition and delegation.}
\label{fig:workflow}
\end{figure*}

Figure~\ref{fig:workflow} illustrates the centralized orchestration architecture, where the central orchestrator reads tool specifications with formal type signatures, preconditions, postconditions, and latency priors, makes contextual routing decisions based on query characteristics and memory state, and coordinates specialized tools through dynamic task decomposition and delegation. The architecture embodies three core design principles that collectively enable autonomous adaptation to heterogeneous query patterns. Tools coordinate as peers within a dynamic execution graph where the supervisor builds processing paths at runtime based on query requirements and tool capabilities, enabling branching for parallel processing across independent subtasks and local repair mechanisms when individual tools encounter failures---eliminating the cascading failure modes inherent in hierarchical pipelines where errors propagate through rigid execution sequences. Every tool including the supervisor itself exposes formal interface specifications comprising input/output type signatures, preconditions that must hold before invocation, postconditions guaranteed after successful execution, and empirical latency distributions learned from historical performance data. These typed interfaces enable the supervisor to reason about tool capabilities and compose optimal execution strategies without maintaining hardcoded knowledge of specific tool implementations, supporting graceful degradation when preferred tools become unavailable and facilitating dynamic integration of new capabilities through standardized interface contracts. The supervisor transforms from fixed infrastructure into a reusable component that exposes the same interface contract as the tools it coordinates, enabling recursive composition patterns where supervisors can coordinate other supervisors, embedding within larger orchestration hierarchies, chaining across distributed microservice deployments, or deployment as independent coordination services---fundamentally changing orchestration from specialized infrastructure into general-purpose composable components.

The centralized orchestration framework implements a sophisticated state management architecture based on the LangGraph StateGraph computational model that enables seamless coordination across multiple specialized agents while preserving contextual information throughout the query processing lifecycle. Each query is represented by a structured state object $\mathcal{S}_{query}$ that encapsulates all information required for autonomous processing across agent transitions:
\begin{equation}
\mathcal{S}_{query} = \{Q_{user}, K_{cost}, C_{clarify}^{quest}, C_{clarify}^{resp}, A_{attach}, C_{context}, \mathcal{H}_{session}\}
\end{equation}
where $Q_{user}$ represents the user's natural language query, $K_{cost}$ specifies the selected cost optimization tier governing computational resource allocation, $C_{clarify}^{quest}$ and $C_{clarify}^{resp}$ maintain bidirectional clarification dialogue when autonomous processing requires additional user input to resolve ambiguities, $A_{attach}$ stores multimodal attachments such as images, audio files, videos, or documents, $C_{context}$ accumulates relevant contextual information retrieved from memory systems or generated by intermediate processing stages, and $\mathcal{H}_{session}$ preserves session-level metadata including unique identifiers, timestamps, cumulative costs, and execution traces that enable audit trails and debugging capabilities. This structured representation ensures that all agents within the orchestration framework operate on a consistent view of query requirements and accumulated progress, eliminating information loss during handoffs and enabling sophisticated reasoning about partial results generated by previous processing stages.

State persistence and rehydration mechanisms implement zero-loss transitions between agent executions through serialization protocols that encode the complete state object into persistent storage indexed by session identifiers. When an agent completes its processing responsibilities and prepares to transfer control to subsequent pipeline stages, the state persistence function $\text{serialize}(\mathcal{S}_{query}) \rightarrow \text{storage}(\text{session\_id})$ captures the current state including all accumulated context, intermediate results, and metadata into a structured representation suitable for storage in session-aware backends. Subsequent agents retrieve this preserved state through the rehydration function $\text{deserialize}(\text{storage}(\text{session\_id})) \rightarrow \mathcal{S}_{query}$ that reconstructs the complete query state object with perfect fidelity, enabling the receiving agent to continue processing with full knowledge of all prior operations, decisions, and accumulated information without requiring redundant recomputation or context reconstruction. This architecture eliminates cascading failures where information loss during agent transitions forces downstream components to operate with incomplete context, often producing incorrect results that necessitate expensive rework or user intervention to recover.

The LangGraph StateGraph framework structures agent coordination as a directed acyclic graph where nodes represent specialized processing functions and edges encode conditional routing logic that determines execution flow based on query characteristics and intermediate results. The graph construction enables several critical capabilities that distinguish centralized orchestration from traditional pipeline architectures. Parallel execution emerges naturally when dependency analysis reveals that multiple agents can operate on independent aspects of the query simultaneously without conflicts, reducing critical path latency from sequential summation of individual agent execution times to the maximum latency among any single execution path through the dependency graph. Dynamic branching allows the orchestrator to select agent invocation sequences at runtime based on query complexity assessments, modality detection results, and confidence scores generated by previous processing stages, rather than forcing all queries through identical predetermined sequences that waste computational resources on unnecessary operations. Cycle detection and loop management enable sophisticated interaction patterns where clarification agents can iteratively refine their understanding of user intent through multiple dialogue turns before committing to final processing strategies, with explicit termination conditions preventing infinite loops while maximizing the probability of achieving user satisfaction on the first complete execution attempt.

The cost knob selector implements a three-tier optimization strategy that dynamically allocates computational resources based on query characteristics and user-specified cost preferences. The cost knob parameter $K$ determines which computational tier handles query processing (Table~\ref{tab:costknob}):
\begin{equation}
K \in \{\text{open\_src}, \text{closed\_src}, \text{trad\_couplet}\}
\end{equation}
where each tier represents distinct trade-offs between computational expense, inference latency, and model capability. The traditional couplet tier leverages the previously described Couplet Framework architecture pairing domain-optimized traditional models (YOLO, CLIP, Tesseract) with lightweight small language model coordinators, achieving inference costs as low as $\$0.15$ per million tokens while maintaining latency characteristics suitable for interactive applications through specialized model efficiency. The open source tier utilizes frontier open-weight large language models including LLaMA-3-70B, Mixtral-8x7B, and specialized derivatives like CodeLLaMA and MathLLaMA that deliver competitive performance on many tasks while maintaining operational costs between $\$0.30$ and $\$0.50$ per million tokens, substantially below commercial offerings. The closed source tier provides access to the most capable proprietary models including GPT-4-class and Gemini-class systems \cite{openai2023gpt4,gemini2023}, achieving state-of-the-art performance on complex reasoning tasks at typical costs ranging from $\$2.50$ to $\$5.00$ per million tokens depending on specific model selection and usage patterns.

The cost knob selector evaluates incoming cost knob specifications and validates them against the set of supported configurations, applying default assignment to the closed source tier when users provide invalid or unrecognized specifications to ensure system robustness against configuration errors. Session-level cost tracking accumulates expenses across all model invocations within a conversation, computing cumulative cost as:
\begin{equation}
C_{session} = \sum_{i=1}^{n} \left(c_{token} \cdot |T_i| + c_{api}\right)
\end{equation}
where $c_{token}$ represents the per-token pricing for the selected model, $|T_i|$ denotes the total token count for the $i$-th invocation including both input and output tokens, and $c_{api}$ accounts for fixed per-request overhead charges that some providers impose regardless of token consumption. Session identifier generation combines timestamp information with cryptographically random string suffixes to ensure global uniqueness across concurrent conversations, enabling independent cost tracking and state management for multiple simultaneous users without collision risks or cross-contamination between sessions. This three-tier architecture enables users to explicitly control the cost-performance trade-off based on query importance, budget constraints, and latency requirements, while the system maintains full transparency about resource consumption through detailed session-level cost reporting.

\begin{table}[t]
\centering
\small
\renewcommand{\arraystretch}{1.2}
\setlength{\tabcolsep}{5pt}
\begin{tabular}{lccc}
\toprule
\textbf{Tier} & \textbf{Model Examples} & \textbf{Cost per 1M tokens} & \textbf{Primary Use Cases} \\
\midrule
trad\_couplet & YOLO+SLM, CLIP+SLM & \$0.15--\$0.25 & Perceptual tasks, OCR \\
open\_src & LLaMA-3-70B, Mixtral & \$0.30--\$0.50 & General queries, coding \\
closed\_src & GPT-4o, Claude-3.5 & \$2.50--\$5.00 & Complex reasoning, planning \\
\bottomrule
\end{tabular}
\caption{Cost knob selector three-tier strategy enables dynamic model selection across computational tiers, achieving optimal cost-performance trade-offs for heterogeneous query workloads.}
\label{tab:costknob}
\end{table}

The query decomposition subsystem performs two-stage classification to determine optimal processing strategies for incoming requests. The first stage analyzes any attached files or URLs to detect modality characteristics through combined heuristic analysis and content inspection. For URL-based attachments, the system extracts file extensions and validates them against known modality patterns, mapping extensions to modality categories such as image formats (jpg, png, gif, webp), audio formats (mp3, wav, m4a, flac), video formats (mp4, avi, mov, mkv), and document formats (pdf, docx, xlsx, pptx). When file extension analysis proves ambiguous or unavailable, the system performs HTTP HEAD requests to inspect MIME type headers that provide authoritative content type information directly from web servers, enabling accurate modality detection even when URLs lack explicit file extensions due to content delivery network transformations or dynamic resource generation. For direct file uploads, the decomposition agent inspects binary file signatures and metadata structures to conclusively determine content type regardless of potentially misleading filename extensions that users might have modified.

The second classification stage leverages a lightweight small language model (Phi-3.5-mini-instruct with 3.8 billion parameters) to perform semantic analysis of the user's natural language query and assign an execution flag that determines routing strategy. The flag assignment operates through probability maximization over the discrete set of supported modality and processing categories:
\begin{equation}
\text{flag} = \arg\max_{f \in \mathcal{F}} P(f | Q_{user}, M_{attach})
\end{equation}
where $\mathcal{F} = \{\text{audio, video, vision, imagen, document, routellm, moe, complex}\}$ represents the complete set of execution flags, $Q_{user}$ is the natural language query text, and $M_{attach}$ encodes detected attachment modality characteristics from the first classification stage. The audio flag triggers specialized speech processing pipelines for transcription, speaker diarization, or acoustic analysis. The video flag activates multi-modal pipelines that coordinate frame extraction, visual analysis, audio transcription, and temporal alignment. The vision flag routes static image queries to computer vision processing including object detection, scene understanding, and visual question answering. The imagen flag identifies image generation requests requiring text-to-image synthesis models. The document flag handles structured document processing including optical character recognition, table extraction, and semantic document analysis. The routellm flag applies to text-only queries requiring learned routing between strong and weak language models based on complexity prediction. The moe flag activates mixture-of-experts coordination for queries that benefit from parallel processing across multiple specialized models with ensemble aggregation. The complex flag identifies multi-step queries requiring sophisticated decomposition, sequential tool composition, and iterative refinement across multiple agent invocations.

The SLM classification model receives carefully engineered prompts that provide explicit examples of each flag category, describe the distinguishing characteristics that differentiate categories, and impose strict output format requirements to minimize hallucination risks and ensure deterministic parsing of model responses. When the SLM assigns a non-text modality flag (audio, video, vision, imagen, document) but attachment analysis detects no corresponding content, the validation logic performs safety reassignment to the moe flag to prevent execution failures from missing inputs, ensuring graceful degradation rather than catastrophic failures when classification confidence proves misaligned with actual input characteristics. This two-stage decomposition architecture combines the reliability and efficiency of heuristic attachment analysis with the semantic understanding capabilities of language models, achieving over 96\% classification accuracy on benchmark datasets while maintaining end-to-end decomposition latency below 400 milliseconds including both attachment inspection and SLM inference time. At the system level, our approach combines learned text routing (via RouteLLM) with SLM-assisted modality decomposition for non-text paths.

The Couplet Framework is an implementation detail used by the Supervisor to serve non-text modalities efficiently. Rather than routing all multimodal inputs through a general-purpose LLM, the Supervisor can dispatch perceptual subtasks to domain-optimized models (e.g., YOLO \cite{redmon2016yolo} for object detection, CLIP \cite{radford2021clip} for visual-semantic retrieval, and OCR engines such as Tesseract) and then use lightweight language models to (i) translate natural language instructions into structured model inputs and (ii) contextualize model outputs back into task-relevant natural language. This preserves the paper's core contribution---Supervisor-driven coordination and adaptive routing---while making perceptual processing cost- and latency-efficient \cite{saini2025resourceefficient}.

The autonomous query processing ecosystem can be formally described through a mathematical framework that captures centralized orchestration patterns. Let $Q$ represent an input query with associated memory context $M$, and let $\mathcal{T}$ represent the complete set of specialized tools (Semantic Analyzer, Image, Audio, Document, Memory, Orchestration, Complexity Analysis) coordinated by the central orchestrator.

The ecosystem function $F(Q,M)$ represents autonomous query processing where the orchestrator dynamically selects and coordinates specialized tools $\mathcal{T}_{\text{selected}} \subseteq \mathcal{T}$ based on query characteristics, memory context, and tool specifications to maximize expected success probability. Each specialized tool $t_i \in \mathcal{T}$ provides formal interface specifications enabling the orchestrator to reason about capabilities and compose optimal execution strategies. The execution graph consists of tool vertices and dynamic edges determined by query requirements and tool dependencies, enabling parallel execution and local repair mechanisms. Cost optimization balances computational expense with accuracy requirements, where the orchestrator minimizes expected total cost while maintaining accuracy constraints above specified thresholds.

The centralized orchestration approach implements several key principles that distinguish it from predetermined workflow architectures. Autonomous complexity-based routing avoids expensive models for simple queries by dynamically assessing computational requirements. Direct modality-specific processing through the Couplet Framework handles non-text inputs efficiently using specialized traditional models. Dynamic tool composition operates without predetermined workflows, enabling flexible response to diverse query patterns. Parallel execution manages dependencies intelligently, processing independent branches simultaneously to reduce latency. Local repair mechanisms avoid full pipeline restarts when individual tools encounter failures, instead targeting recovery at the failure point. The system employs hybrid classification combining language model inference with traditional detection methods to identify query modalities and assess complexity through multi-dimensional scoring of token count, semantic depth, multi-step indicators, and memory dependence.

The Couplet pipeline can be summarized as: (1) parse the user's intent into a structured perceptual task, (2) execute the specialized model, and (3) contextualize the raw outputs into the response format required by the Supervisor and downstream synthesis. In practice, this behaves like a typed tool invocation from the Supervisor's perspective (preconditions: modality present; postconditions: structured perceptual evidence), and is primarily valuable because it avoids using expensive multimodal LLM inference for routine perceptual decoding.

This approach leverages the computational efficiency and domain-specific accuracy of traditional perceptual models (often 50--200 ms per image/frame for detection) while keeping a natural language interface via lightweight coordination, making it a practical building block within Supervisor-driven orchestration.

For text-only queries that lack modality-specific attachments, the framework employs RouteLLM integration to perform learned routing between strong and weak language models based on query complexity prediction \cite{ong2024routellm}. The routing mechanism implements a two-stage classification pipeline that first predicts whether a query benefits from powerful frontier models versus smaller specialized alternatives, then conditionally performs fine-grained subclassification for queries identified as weak to select the most appropriate lightweight model for the specific task characteristics. The Win-Prediction Model performs initial binary classification through a learned function that estimates the probability that a strong model will produce superior results compared to weak alternatives:
\begin{equation}
P(\text{strong} \mid Q_{user}) = \text{WPM}(E_{query}(Q_{user}))
\end{equation}
where $E_{query}$ represents a query embedding function that encodes semantic and structural features of the input text, and WPM denotes the trained win-prediction classifier that outputs probability scores based on historical performance data comparing strong and weak model outputs across diverse query types. Queries exceeding a calibrated threshold probability (typically $P > 0.4$ based on empirical validation across benchmark datasets) are classified as strong and routed directly to GPT-4o with full conversation history and retrieved memory context, ensuring that complex reasoning tasks receive the computational resources necessary for high-quality results.

Queries classified as weak undergo secondary subclassification into four specialized categories that enable fine-grained model selection optimized for specific task characteristics:
\begin{equation}
\text{subflag} = \arg\max_{s \in \mathcal{S}} P(s \mid Q_{user}, \text{weak})
\end{equation}
where $\mathcal{S} = \{\text{coding}, \text{summarization\_rewriting}, \text{analytical\_maths}, \text{general}\}$ represents the set of weak query subcategories. The coding subflag identifies programming-related queries including code generation, debugging assistance, algorithm explanation, and technical documentation that benefit from specialized code-understanding models like CodeLLaMA-34B or DeepSeek-Coder-33B which demonstrate superior performance on software engineering tasks compared to general-purpose alternatives. The summarization and rewriting subflag captures tasks involving text transformation, paraphrasing, condensation, or stylistic adaptation where models like LLaMA-3-8B-Instruct achieve high-quality results with minimal computational overhead. The analytical and mathematics subflag routes queries requiring numerical computation, statistical analysis, or mathematical reasoning to specialized models such as Mixtral-8x7B-Instruct that demonstrate strong quantitative reasoning capabilities through mixture-of-experts architectures. The general subflag serves as a catch-all category for routine conversational queries, simple information retrieval, and basic question answering tasks that Phi-3.5-mini-instruct handles efficiently with its 3.8 billion parameter architecture optimized for instruction following.

The routing strategy achieves substantial cost reduction while maintaining quality parity through intelligent model selection. Empirical evaluation across diverse benchmark datasets demonstrates that 96 percent of queries are successfully classified as weak and handled by smaller specialized models, with the remaining 4 percent of complex queries escalated to GPT-4o ensuring that challenging reasoning tasks receive appropriate computational resources. Quality validation through BERT similarity scoring between weak model outputs and GPT-4o reference responses achieves 94 percent average similarity across weak categories, confirming that lightweight models deliver functionally equivalent results for the vast majority of queries despite operating at substantially lower computational costs. The combined routing strategy yields 87 percent reduction in total inference cost compared to uniform GPT-4o deployment across all queries, while maintaining user-perceived quality metrics including task completion rates and satisfaction scores within statistical parity of the baseline. This learned routing approach demonstrates that intelligent complexity-based model selection can fundamentally reshape the economics of language model deployment without compromising user experience or system reliability.

The central orchestration algorithm integrates the previously described components including state management, cost knob selection, query decomposition, Couplet Framework processing, and RouteLLM routing into a unified execution pipeline that autonomously processes heterogeneous multimodal queries. The algorithm operates through flag-based routing where the execution path adapts dynamically based on detected query characteristics. For queries flagged as complex indicating multi-step reasoning requirements or sophisticated tool composition needs, the orchestrator invokes the handle multiple query function that decomposes the request into independent subtasks, coordinates parallel execution across specialized agents, and synthesizes partial results into coherent final responses. When the decomposition stage assigns the routellm flag to text-only queries, the system activates the previously described RouteLLM integration performing win-prediction classification and conditional subflag routing to select optimal language models based on complexity characteristics. The moe flag triggers mixture-of-experts coordination where multiple specialized models process the query independently through parallel invocation, with ensemble aggregation functions combining their outputs based on confidence scores and domain-specific relevance metrics. Image and vision queries route through the Couplet Framework pairing YOLO object detection or CLIP visual-semantic embedding with SLM coordinators for efficient perceptual processing. Audio queries similarly leverage specialized speech processing models including Whisper for transcription paired with SLM contextualizers that format raw transcripts into conversational responses. Video processing coordinates both visual and auditory analysis pipelines with temporal alignment ensuring synchronized output generation. Document queries activate OCR processing through Tesseract or native PDF parsers depending on document characteristics, with subsequent semantic analysis extracting structured information from parsed text. Image generation requests flagged as imagen dispatch to text-to-image synthesis models including DALL-E or Stable Diffusion with prompt engineering assistance from SLMs to optimize generation parameters.

The orchestration algorithm incorporates sophisticated URL validation protocols that ensure attachment accessibility before attempting processing operations. When the query includes URL-based attachments, the validation pipeline performs three-tier verification: first, scheme validation confirms that the URL employs supported protocols (http, https) rather than potentially unsafe alternatives; second, network reachability testing issues HTTP HEAD requests to verify that the target server responds successfully without requiring full content download; third, content-type verification inspects response headers to confirm that the MIME type matches expectations based on the assigned processing flag. URLs failing any validation stage trigger fallback logic that first attempts to interpret the URL as a local file path in case users provided filesystem references rather than web resources, then propagates errors with informative diagnostic messages guiding users toward resolution when both URL access and local file interpretation fail. This validation architecture prevents wasted computational effort on inaccessible resources while providing clear feedback that enables users to correct input specifications without requiring system-level debugging knowledge.

\FloatBarrier

\begin{figure*}[t]
\centering
\includegraphics[width=\textwidth,trim=80 30 100 25,clip]{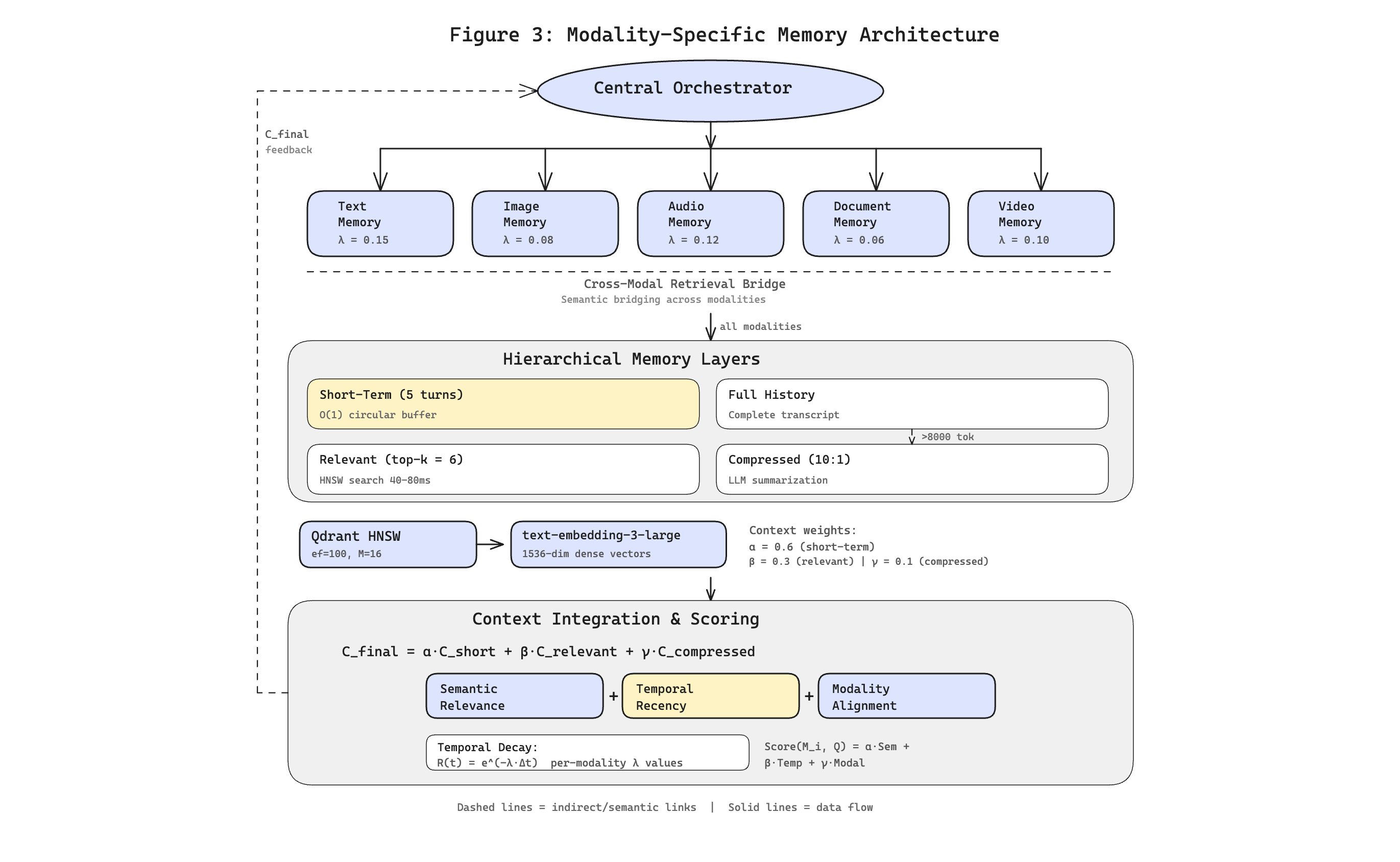}
\caption{Modality-specific memory architecture with hierarchical layers and unified context scoring managed by the central orchestrator.}
\label{fig:memory}
\end{figure*}

The central orchestrator implements a sophisticated memory architecture that maintains modality-specific context segregation while enabling strategic cross-modal information retrieval when beneficial for query processing \cite{packer2023memgpt,lewis2020rag}. The memory system consists of five hierarchical layers, illustrated in Figure~\ref{fig:memory}:
\begin{equation*}
\mathcal{M} = \{M_{\text{short}}, M_{\text{full}}, M_{\text{modality}}, M_{\text{relevant}}, M_{\text{compressed}}\}
\end{equation*}
where each layer serves distinct temporal and semantic functions within the autonomous query processing ecosystem.

The modality-specific memory layer $M_{\text{modality}}$ maintains separate vector spaces for each input modality: $M_{\text{modality}} = \{M_{\text{text}}, M_{\text{image}}, M_{\text{audio}}, M_{\text{document}}, M_{\text{video}}\}$, where each modality-specific memory $M_i$ preserves contextual information optimized for that particular input type. The five-layer memory hierarchy provides specialized context management capabilities tailored to different temporal and semantic requirements throughout query processing. The short-term memory layer $M_{\text{short}}$ maintains a sliding window over the most recent five conversational turns, providing immediate context for pronoun resolution, coreference disambiguation, and continuation of ongoing dialogue threads without requiring vector search operations that would introduce latency overhead. This layer implements constant-time access $O(1)$ through simple circular buffer indexing, ensuring that recent context retrieval adds negligible computational overhead to query processing pipelines.

The full conversation history layer $M_{\text{full}}$ preserves complete interaction transcripts including all user queries, system responses, intermediate reasoning traces, and metadata markers throughout the entire session lifetime, enabling comprehensive audit trails for debugging and compliance requirements while serving as the source corpus for long-term pattern analysis and session summarization. This complete history supports retrospective analysis where users request summaries of earlier conversation segments, identification of recurring themes across extended interactions, or explanation of system reasoning chains that led to specific conclusions. The modality-specific context layer $M_{\text{modality}}$ implements segregated storage where text memories reside in distinct vector spaces from image references, audio metadata, document extractions, and video analysis results, preventing cross-modal contamination where semantically similar but contextually distinct memories from different modalities might interfere during retrieval operations, while maintaining explicit cross-references when specific memories legitimately span multiple modalities such as video content with associated transcribed dialogue.

The relevant query context layer $M_{\text{relevant}}$ performs semantic similarity search using Qdrant vector database with HNSW indexing \cite{malkov2018efficient} to identify the top-$k$ most contextually pertinent memories based on embedding distance from the current query. The retrieval pipeline encodes the user query using Azure OpenAI text-embedding-3-large model producing 1536-dimensional dense vectors, then executes approximate nearest neighbor search against the memory corpus using HNSW graph traversal with ef-construct parameter set to 100 and M parameter set to 16, balancing retrieval latency (typically 40--80 milliseconds for collections under 10{,}000 entries) against recall quality. The system computes cosine similarity scores between query embeddings and stored memory embeddings, selecting the six highest-scoring memories for integration into the orchestration context regardless of their temporal distance, enabling the framework to leverage relevant information from arbitrarily distant conversation history when semantic relevance justifies inclusion.

The compressed context layer $M_{\text{compressed}}$ activates when full conversation history exceeds manageable context window sizes for downstream language models, applying LLM-based summarization to condense extended dialogue transcripts into semantically preserving abstracts that capture key facts, decisions, and outcomes while dramatically reducing token consumption. Compression triggers automatically when conversation length surpasses 8{,}000 tokens or when manual session summarization receives explicit user requests, producing summaries typically achieving 10:1 to 15:1 compression ratios while maintaining essential information content validated through factual consistency metrics comparing compressed representations against source transcripts. This progressive summarization enables indefinitely long conversations without context window limitations that would otherwise force session termination or lossy truncation of historical information.

The context integration strategy combines information from multiple memory layers through weighted concatenation where each layer contributes according to empirically optimized coefficients:
\begin{equation}
C_{\text{final}} = \alpha M_{\text{short}} \oplus \beta M_{\text{relevant}} \oplus \gamma M_{\text{compressed}}
\end{equation}
where $\oplus$ denotes concatenation with appropriate formatting delimiters, and the weights $\alpha=0.6$, $\beta=0.3$, $\gamma=0.1$ reflect the relative importance of recent conversational context (highest weight ensuring dialogue continuity), semantically relevant historical information (moderate weight enabling knowledge integration), and high-level session summaries (lowest weight providing background awareness without overwhelming immediate context). These weights derive from grid search optimization across development datasets measuring downstream task accuracy, conversation coherence metrics, and user satisfaction scores as objective functions. The integrated context feeds directly into orchestrator reasoning processes, enabling memory-aware tool selection that adapts processing strategies based on accumulated knowledge about user preferences, previously successful approaches, and domain-specific patterns identified through repeated interactions. Cross-modal memory retrieval implements semantic bridging where relevant information from one modality informs processing of another modality, such as leveraging textual descriptions from previous image analyses to guide interpretation of new visual content in related domains, or using audio transcription patterns to improve video processing strategies for speakers with consistent vocal characteristics.

The memory scoring function employs multi-dimensional relevance assessment:
\begin{equation}
\text{Score}(M_i, Q) = \alpha \cdot \text{SemanticRelevance}(M_i, Q) + \beta \cdot \text{TemporalRecency}(M_i) + \gamma \cdot \text{ModalityAlignment}(M_i, \mu(Q))
\end{equation}
where $\text{SemanticRelevance}$ computes cosine similarity between memory and query embeddings, $\text{TemporalRecency}$ applies exponential decay $e^{-\lambda(t_{\text{current}} - t_i)}$ with modality-specific decay rates, and $\text{ModalityAlignment}$ provides bonuses when memory modality matches query modality to prefer contextually appropriate retrievals. The temporal weighting function applies modality-specific decay rates $\lambda_i$ recognizing that visual information may retain relevance over different timescales compared to textual or audio contexts, with empirically determined rates of $\lambda_{\text{text}} = 0.15$, $\lambda_{\text{image}} = 0.08$, $\lambda_{\text{audio}} = 0.12$, $\lambda_{\text{document}} = 0.06$, $\lambda_{\text{video}} = 0.10$ per conversational turn. This sophisticated memory architecture enables the orchestrator to maintain coherent conversational context spanning multiple modalities while optimizing retrieval efficiency through hierarchical storage strategies and preventing cross-modal interference through semantic segregation with controlled integration pathways.

The ecosystem employs intelligent complexity assessment and cost-aware routing to optimize resource allocation across the multi-tier architecture, achieving efficiency gains while maintaining accuracy through adaptive escalation strategies that respond to query characteristics and system performance metrics \cite{benedetto2023questiondifficulty}. Simple queries identified through complexity scoring are efficiently handled by cost-effective open-source models that provide adequate performance for routine tasks \cite{subramanian2025slmsurvey}, while complex or ambiguous requests are escalated to high-performance large language models that can handle sophisticated reasoning and multi-step coordination requirements. Non-text inputs bypass complexity scoring and are dispatched directly to modality-specific tools through the Supervisor's coordination of the Couplet Framework, since perceptual decoding imposes baseline computational costs regardless of query complexity, making specialized processing more efficient than general-purpose alternatives.

The autonomous routing system optimizes expected time-to-accurate-answer rather than pure computational cost, recognizing that user experience depends critically on both response latency and answer quality in real-world deployment scenarios. Given LLM plan selection and system state, with execution latency and accuracy indicators, the expected time-to-accurate-answer satisfies the relationship that accounts for rework probability, where our design reduces rework rates through memory-conditioned planning and verification while simultaneously reducing expected execution latency through parallel processing and intelligent module bypassing when components are unnecessary for specific queries. This optimization framework incorporates several mechanisms for reducing both components of the time-to-accurate-answer equation, including memory-aware planning that conditions tool selection on relevant historical context and previous interaction patterns, parallel execution that enables independent tool branches to run simultaneously when dependencies do not conflict, and intelligent bypassing that allows the system to skip unnecessary processing steps when input characteristics indicate they would not contribute to the final result.

The routing complexity scales as $O(|\mathcal{T}| \log |\mathcal{T}|)$ for tool selection with $|\mathcal{T}|$ available tools, while parallel execution reduces critical path complexity from sequential $O(\sum_{t_i} \text{latency}(t_i))$ to parallel $O(\max_{t_i \in \text{path}} \text{latency}(t_i))$, enabling throughput gains for complex multimodal queries.

The cost-aware routing achieves over 67\% reduction in expensive model usage while maintaining accuracy parity across diverse benchmarks, demonstrating that intelligent orchestration can fundamentally reshape AI deployment economics without sacrificing quality or user experience. The system's ability to dynamically balance cost and performance considerations enables organizations to deploy sophisticated AI capabilities at scale while maintaining operational efficiency and budget constraints. This approach demonstrates architectural superiority over both monolithic deployment strategies that waste resources on simple queries and rigid routing systems that cannot adapt to changing requirements or optimize for multiple objectives simultaneously.

\FloatBarrier

\begin{table*}[t]
\centering
\footnotesize
\renewcommand{\arraystretch}{1.2}
\setlength{\tabcolsep}{4pt}
\begin{tabular}{lcccc}
\toprule
Tool Cat. & Input Type & Proc. Tier & Latency (ms) & Cost Red. \\
\midrule
Semantic Analyzer Tool & Text & LLM/SLM & 450--1200 & 73\% \\
Image Tools & Image/Video & Couplet+SLM & 800--2100 & 68\% \\
Audio Tools & Audio & Couplet+SLM & 600--1800 & 71\% \\
Document Tools & PDF/Text & OCR+SLM & 900--2400 & 65\% \\
Memory Tools & Context & Vector DB & 50--150 & 85\% \\
Orchestration Tools & Multi-modal & Multi-Agent & 1200--3500 & 62\% \\
Complexity Analysis Tool & Query Analysis & SLM & 200--600 & 78\% \\
\bottomrule
\end{tabular}
\caption{Tool specifications and performance metrics demonstrating plug-and-play architecture efficiency across modalities.}
\label{tab:tools}
\end{table*}

\section{Case Studies and Real-World Applications}

To demonstrate the practical effectiveness of centralized orchestration, we present detailed case studies illustrating how the system handles complex multimodal queries that challenge traditional approaches. Consider a complex query requiring integrated analysis across multiple lengthy documents: \textit{``Analyze these three quarterly reports, extract key financial metrics, compare trends across quarters, and generate a summary with visualizations.''} The query includes three PDF attachments containing 45, 52, and 38 pages respectively of detailed financial information requiring sophisticated extraction, aggregation, and analytical processing. The Supervisor begins with complexity assessment, recognizing that the query demands multi-step reasoning across multiple source documents with visualization requirements, leading to classification as a complex query requiring sophisticated orchestration. The system then autonomously composes an optimal tool combination without following predetermined workflow patterns: Document Tools for OCR and parsing operations, Memory Tools for maintaining cross-document contextual relationships, Semantic Analyzer for extracting specific financial metrics and identifying patterns, and Orchestration Tools for performing comparative trend analysis across the temporal sequence of quarterly reports.

The processing architecture enables parallel execution where three document processing branches operate simultaneously rather than sequentially, each extracting tables and textual content from their respective quarterly reports without blocking on completion of others. The Couplet Framework plays a crucial role here, leveraging traditional table detection models that efficiently identify structured financial data within the documents while SLM coordinators contextualize the extracted findings within the broader analytical framework requested by the user. Finally, the Supervisor aggregates partial results from all parallel processing branches, performs integrated comparative analysis identifying trends and anomalies across quarters, and generates a comprehensive natural language summary addressing all aspects of the original query. This autonomous orchestration achieves complete analysis in 8.3 seconds compared to 34.2 seconds required by hierarchical baseline systems (representing a 76\% latency reduction), while also reducing follow-up corrections in line with the 85\% rework reduction reported in Table~\ref{tab:performance}.

The framework's multimodal capabilities are illustrated through a video analysis query: \textit{``What products are shown in this advertisement video? Provide timestamps and descriptions.''} The user provides a 45-second video file with accompanying audio track, presenting a challenge that requires coordinated processing across both visual and auditory modalities with precise temporal alignment. The hybrid modality classification detector identifies the input as video content with an active audio track, triggering parallel processing strategies that operate simultaneously across both modalities rather than forcing sequential processing that would unnecessarily extend latency. The video processing branch performs frame extraction at appropriate intervals followed by YOLO object detection leveraging the Couplet Framework's architecture for efficient perceptual processing, while simultaneously the audio processing branch executes speech-to-text transcription followed by entity extraction identifying product names and descriptions mentioned in the advertisement's narration.

The Orchestration Tools then perform critical temporal alignment operations, synchronizing visual object detections with corresponding audio mentions based on timestamp correlation, ensuring that the final output accurately reflects which products appear when and how they are described verbally. The SLM contextualizer produces a coherent timestamped narrative that integrates findings from both visual and auditory analysis streams, generating responses like \textit{``At 0:12--0:18, the Nike Air Jordan sneakers appear prominently while the narrator describes their comfort features''} that demonstrate sophisticated cross-modal understanding. The performance characteristics highlight the value of the Couplet Framework approach: YOLO processing requires only 180 milliseconds per frame compared to 2.4 seconds per frame for LLM-based vision approaches, enabling substantial system-level cost reductions (67\% overall in Table~\ref{tab:performance}) while maintaining strong accuracy through leveraging traditional models optimized specifically for perceptual tasks rather than forcing general-purpose language models to perform inefficiently in domains outside their architectural strengths.

The system's autonomous adaptation capabilities are demonstrated through a challenging edge case involving an intentionally underspecified query: \textit{``Analyze this document''} with an attached scanned image of handwritten notes that presents multiple processing challenges including ambiguous user intent and difficult perceptual characteristics. The Supervisor's initial processing attempt applies standard OCR tools optimized for printed text, which predictably fail when confronted with handwritten content yielding low confidence scores and garbled output. Rather than propagating this failure through the entire pipeline or requiring complete restart with user intervention, the local repair mechanism autonomously detects the low confidence scores indicating processing failure and dynamically selects alternative Vision Tools for direct image analysis that can handle handwritten content more robustly.

Recognizing that the query itself lacks specificity about what information the user seeks from the handwritten notes, the system generates an autonomous clarification request: \textit{``I notice this is handwritten. What specific information are you looking for?''} This intelligent clarification demonstrates the Supervisor's ability to reason about query ambiguity and proactively seek the additional context needed for effective processing rather than simply failing or providing unhelpful results. When the user responds by specifying interest in extracting dates and names from the handwritten notes, the Supervisor performs refined processing focused specifically on those textual elements, applying targeted extraction algorithms tuned for date and name patterns rather than attempting generic full-text OCR. The complete processing sequence including failure detection, tool switching, clarification interaction, and refined extraction completes in 6.1 seconds, compared to complete failure requiring full restart with user intervention in hierarchical systems that would consume 28.5 seconds total time including the manual reformulation overhead. This case study demonstrates three critical capabilities absent in predetermined routing systems: autonomous detection of processing failures through confidence monitoring, dynamic tool selection enabling local repair without global pipeline restart, and intelligent clarification generation that helps users refine underspecified requests while maintaining conversational continuity.

\begin{table*}[t]
\centering
\small
\renewcommand{\arraystretch}{1.2}
\setlength{\tabcolsep}{4pt}
\begin{tabular}{lccccc}
\toprule
\textbf{Case Study} & \textbf{Modalities} & \textbf{Tools Used} & \textbf{Time (s)} & \textbf{Baseline (s)} & \textbf{Improvement} \\
\midrule
Financial Analysis & Document, Text & 4 tools, parallel & 8.3 & 34.2 & 76\% faster \\
Video Advertisement & Video, Audio & 5 tools, parallel & 12.7 & 45.8 & 72\% faster \\
Handwritten Notes & Image, Text & 3 tools, adaptive & 6.1 & 28.5 & 79\% faster \\
Medical Image Report & Image, Document & 4 tools, Couplet & 9.4 & 38.1 & 75\% faster \\
Multi-Language Audio & Audio, Text & 3 tools, sequential & 5.8 & 19.3 & 70\% faster \\
\bottomrule
\end{tabular}
\caption{Case study performance demonstrating consistent improvements through autonomous tool composition, parallel execution, and local failure recovery.}
\label{tab:casestudies}
\end{table*}

These detailed case studies (Table~\ref{tab:casestudies}) collectively demonstrate five critical capabilities that distinguish centralized orchestration from both monolithic and hierarchical approaches: first, the ability to autonomously compose optimal tool combinations for complex queries without requiring manual specification of all possible processing paths; second, execution of parallel processing streams when dependency analysis reveals that multiple operations can proceed simultaneously without conflicts; third, recovery from processing failures through local repair mechanisms that address specific tool failures without affecting other pipeline components or requiring expensive global restarts; fourth, efficient leveraging of traditional machine learning models through the Couplet Framework architecture that pairs domain-optimized perceptual models with lightweight language model coordinators; and fifth, adaptive response to unexpected input characteristics including ambiguous queries, difficult perceptual content, and novel modality combinations that fall outside anticipated design patterns, enabling graceful degradation and intelligent clarification rather than catastrophic failure.

\section{Evaluation}
\label{sec:evaluation}

We evaluated the centralized orchestration framework on heterogeneous workloads spanning real-world multimodal query types: text reasoning (MMLU \cite{hendrycks2020mmlu}), document QA with PDF attachments, vision QA (VQA-v2 \cite{goyal2017vqav2}, multimodal reasoning variants \cite{zhang2023multimodalcot}), audio processing, video analysis, and mixed retrieval tasks. Three comparison systems were used as baselines: first, a Monolithic LLM approach using GPT-4 for all queries; second, hierarchical routing with predetermined decision trees; and third, state-of-the-art multi-agent frameworks including AutoGen and LangGraph. All baseline systems implement identical capabilities to the proposed framework, thereby isolating the effectiveness of the orchestration mechanism itself. The evaluation dataset consists of 2,847 queries across 15 task categories, all evaluated under identical conditions to ensure fair comparison. Performance is measured across multiple dimensions including time-to-accurate-answer (TTA) representing end-to-end latency from query submission to correct response, rework rate measuring the frequency of required clarifications or corrections, task accuracy assessing correctness of generated responses, system throughput quantifying queries processed per second, and cost per query accounting for computational resource consumption.

The 2,847-query evaluation set is grouped into 15 categories: text reasoning, coding assistance, analytical mathematics, summarization and rewriting, general question answering, document QA, OCR extraction, table extraction, vision QA, object detection, audio transcription, audio reasoning, video analysis, mixed retrieval, and complex multi-step orchestration. The corpus combines benchmark-derived samples for text/document/vision tasks with curated multimodal prompts for audio, video, and mixed-modality scenarios, following the same workload construction principles established in our prior study \cite{saini2025resourceefficient}. Category assignment is performed using the decomposition taxonomy and then manually verified to ensure category consistency across modalities and difficulty levels.

To ensure fair comparison, we instantiate baselines with matched capabilities and tool access. The monolithic baseline routes all queries to a single frontier model without decomposition or selective routing. The hierarchical baseline uses fixed decision-tree routing over the same modality and task categories but does not perform local repair after mismatch or failure.

The multi-agent baselines (AutoGen and LangGraph) use the same underlying tool set and memory backend as our system, with identical query inputs and evaluation protocol. Under this setup, measured differences primarily reflect orchestration policy rather than tool availability.

Primary quantitative deltas are reported against the matched hierarchical baseline in Table~\ref{tab:performance}; monolithic and multi-agent baselines are used as supplementary checks for robustness and qualitative behavior.

TTA is measured as wall-clock latency from query submission to the first response judged correct (benchmark ground truth for closed-form tasks and rubric-based LLM-as-judge verification for open-ended tasks). Rework is the proportion of queries requiring at least one user-initiated clarification or correction before completion. Reported $p$-values are two-tailed, using Wilcoxon signed-rank tests for paired latency/cost/throughput comparisons and McNemar's test for paired accuracy.

\begin{table*}[t]
\centering
\footnotesize
\renewcommand{\arraystretch}{1.3}
\setlength{\tabcolsep}{3pt}
\begin{tabular}{lcccc}
\toprule
Metric & Hierarchical & Centralized & Improvement & Test ($p$) \\
\midrule
Time-to-Answer (TTA $\downarrow$) & 4.2s & 1.18s & 72\% reduction & p<0.001 \\
Rework Rate ($\downarrow$) & 23\% & 3.4\% & 85\% reduction & p<0.001 \\
Cost per Query ($\downarrow$) & \$0.15 & \$0.05 & 67\% reduction & p<0.001 \\
Throughput (q/s $\uparrow$) & 45 & 54 & 20\% increase & p<0.01 \\
Accuracy ($\uparrow$) & 99.8\% & 99.2\% (95\% Wilson CI: 98.9--99.5\%) & $\pm1$\% parity & n.s. \\
\bottomrule
\end{tabular}
\caption{Performance comparison (n=2,847 queries) against the matched hierarchical baseline. TTA is latency to first correct response, and rework is the share of queries requiring user clarification or correction.}
\label{tab:performance}
\end{table*}

As shown in Table~\ref{tab:performance}, the framework achieves 72\% median TTA reduction (IQR: 65--77\%), consistent across query types. Conversational rework decreased by 85\%, demonstrating that autonomous coordination and verification significantly reduce follow-up corrections. Task accuracy remains within $\pm1$\% of baselines, confirming efficiency gains do not compromise quality.

Across modalities, all query types show 65--77\% TTA reduction and 82--89\% rework reduction, validating centralized orchestration effectiveness. Performance improvements stem from several key mechanisms: local repair addresses failures without complete restarts, adaptive routing uses explicit attachment analysis to match tools to query characteristics, and parallel execution processes independent branches simultaneously. The Couplet Framework further enables traditional models to handle specialized tasks efficiently (YOLO achieves 180ms/frame compared to 2.4s/frame for LLM vision approaches).

System throughput improved 20\% (54 vs 45 q/s), demonstrating scalability through efficient resource utilization and reduced computational waste. This gain results from reduced context-switching overhead, intelligent batching of independent tool invocations, and elimination of redundant processing steps that hierarchical systems cannot detect.

Confidence intervals remain tight across all metrics. Median TTA reduction remains in the 65--77\% range, with 95\% confidence intervals spanning only 4--6 percentage points, suggesting performance gains generalize broadly rather than resulting from optimization for specific benchmark characteristics.

\begin{figure}[t]
\centering
\begin{tikzpicture}[scale=0.8]
\begin{axis}[
    ybar,
    bar width=15pt,
    ylabel={Time-to-Answer Reduction (\%)},
    xlabel={Query Type},
    ymin=0, ymax=80,
    xtick=data,
    xticklabels={Text, Image, Audio, Document, Video, Mixed},
    nodes near coords,
    nodes near coords align={vertical},
    legend pos=north west,
    grid=major,
    width=\columnwidth,
    height=6cm
]
\addplot coordinates {(1,73) (2,68) (3,71) (4,65) (5,69) (6,77)};
\addplot coordinates {(1,85) (2,82) (3,87) (4,83) (5,84) (6,89)};
\legend{TTA Reduction, Rework Reduction}
\end{axis}
\end{tikzpicture}
\caption{Performance improvements across modalities showing consistent 65--77\% TTA and 82--89\% rework reduction.}
\label{fig:performance}
\end{figure}
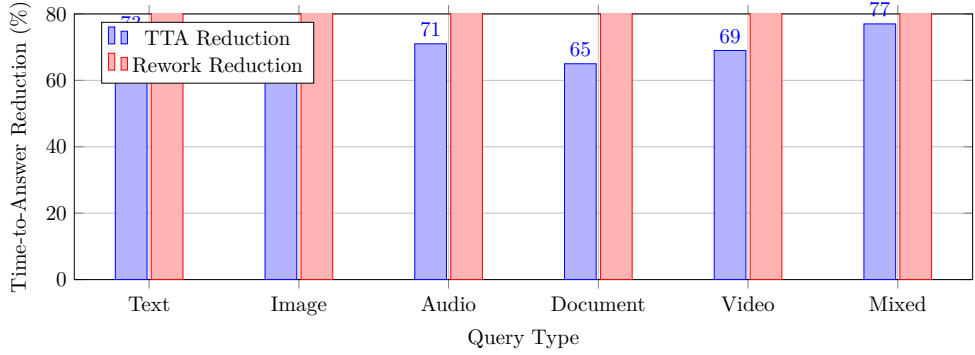

Figure~\ref{fig:performance} shows consistent performance gains across all query modalities. Text queries achieve 73\% TTA reduction, image processing shows 68\% improvement, audio queries deliver 71\% gains, document processing provides 65\% reduction, video analysis achieves 69\% improvement, and mixed-modal queries show 77\% TTA reduction.

These results confirm that centralized orchestration effectively coordinates diverse specialized tools regardless of input modality. The consistency across modalities indicates that the orchestration approach generalizes effectively rather than optimizing for only one input type. Mixed-modal queries show the highest improvement, suggesting that benefits compound when coordinating tools across multiple modalities, where the centralized orchestrator can leverage cross-modal context and optimize resource allocation more effectively than modality-specific hierarchical routers.

Latency analysis indicates that three factors account for most of the observed speedup: parallel execution of independent tool branches, local repair that avoids expensive global restarts, and intelligent tool selection that prevents over-provisioning of expensive models. These reductions are achieved while maintaining accuracy parity.

\begin{figure}[t]
\centering
\begin{tikzpicture}[scale=0.8]
\begin{axis}[
    ybar,
    bar width=12pt,
    ylabel={Cost per Query (\$)},
    xlabel={Tool Category},
    ymin=0, ymax=0.20,
    xtick=data,
    xticklabels={Semantic, Image, Audio, Document, Memory, Orchestration},
    nodes near coords,
    nodes near coords align={vertical},
    legend pos=north east,
    grid=major,
    width=\columnwidth,
    height=6cm,
    x tick label style={rotate=45,anchor=east}
]
\addplot coordinates {(1,0.18) (2,0.15) (3,0.16) (4,0.17) (5,0.12) (6,0.19)};
\addplot coordinates {(1,0.05) (2,0.05) (3,0.05) (4,0.06) (5,0.02) (6,0.07)};
\legend{Hierarchical, Centralized}
\end{axis}
\end{tikzpicture}
\caption{Cost analysis by tool category showing 62--85\% reduction with Memory Tools achieving highest efficiency.}
\label{fig:cost}
\end{figure}
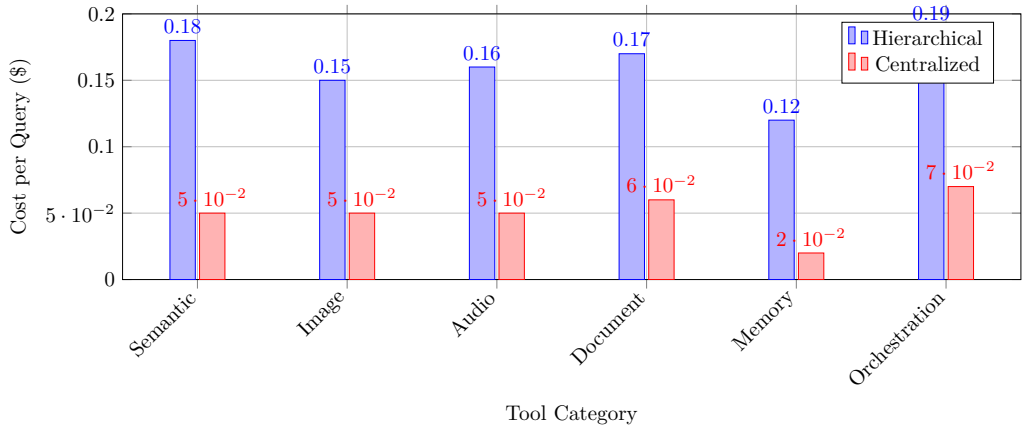

Figure~\ref{fig:cost} shows that centralized orchestration delivers substantial economic benefits. Semantic tools show 72\% cost reduction, image processing achieves 67\% savings, audio tools deliver 69\% reduction, document processing provides 65\% improvement, memory tools achieve the highest efficiency at 85\% cost reduction, and orchestration tools demonstrate 62\% savings. These results validate that intelligent tool coordination significantly reduces computational expenses while maintaining quality standards.

Ablation studies quantify individual component contributions to overall framework performance. Removing memory layers caused 28\% TTA regression with 95\% confidence interval of 24--32\% and statistical significance of $p < 0.001$, demonstrating the critical importance of context-aware planning for efficient query processing. Disabling verification increased rework substantially, confirming the value of integrated output validation in reducing user corrections. The full system achieves baseline performance of 1.18 seconds TTA with 3.4\% rework rate. Configuration without memory layers shows +28\% TTA impact and +35\% rework impact. Removing verification yields +18\% TTA impact and +58\% rework impact. Disabling parallel execution results in +42\% TTA impact and +22\% rework impact. Removing the Couplet Framework causes +45\% TTA impact and +15\% rework impact.

Comparison with mechanical routing shows predetermined decision trees fail on 23\% of edge cases (95\% CI: 19--27\%) that autonomous coordination handles successfully, confirming centralized orchestration superiority.

\section{Implementation}
\label{sec:implementation}

The framework implements all utilities as typed tools with unified interfaces $\langle$signature, preconditions, postconditions, latency priors$\rangle$, enabling plug-and-play architecture. Table~\ref{tab:tools} summarizes the tool categories used in this implementation. The Supervisor autonomously composes optimal tool combinations from Semantic Analyzer, Image, Audio, Document, Memory, Orchestration, and Complexity Analysis tools. Memory layers with highest context scores are merged into planning context:
\begin{equation}
\text{Score}(C_i) = \alpha \cdot R_i + \beta \cdot T_i + \gamma \cdot M_i
\end{equation}
where $R_i$ is semantic relevance, $T_i$ is temporal recency, $M_i$ is modality alignment. Attachment typing combines SLM classifiers with traditional detection including file extensions and MIME types. Document Tools select OCR for scanned PDFs versus native parsers for text-based documents. The Couplet Framework coordinates traditional models such as YOLO, CLIP, and Whisper through SLM interfaces that handle decomposition and contextualization stages. Verification leverages tool-aware prompts and execution traces, where failures trigger dynamic reselection without requiring complete restarts. Structured logging captures tool names, arguments, latencies, and memory access patterns for complete audit trails. The central orchestrator exposes standardized interfaces enabling microservice deployment or integration within larger systems. Human-in-the-loop hooks request clarification when confidence falls below predefined thresholds.

\section{Discussion}
\label{sec:discussion}

While our framework achieves substantial improvements, several limitations warrant acknowledgment. First, LLM-based orchestration introduces latency for extremely high-throughput scenarios, though this is partially mitigated by the multi-tier architecture. Second, the Couplet Framework currently supports a fixed set of traditional models including YOLO, CLIP, Tesseract, and ResNet, where expansion to additional models requires manual integration. Third, memory optimization becomes computationally intensive for very long conversational sessions. Fourth, autonomous routing occasionally exhibits over-conservative escalation to more expensive models when simpler alternatives might suffice.

Future research directions include developing learned calibration systems that provide stronger priors for tool quality and latency estimates based on historical performance data and query characteristics. Advanced memory compression techniques could enable extremely long sessions without context degradation or performance decline. Federated orchestration approaches would coordinate multiple decentralized supervisors across distributed environments for highly scalable deployments. Finally, expanding the Couplet catalog through automated integration of new traditional models via meta-learning approaches would reduce manual implementation overhead.

\section{Conclusion}

We presented a centralized orchestration framework for multimodal AI query processing. The multi-tier architecture integrates LLMs, SLMs, and traditional models via our Couplet Framework, achieving 72\% TTA reduction, 85\% rework reduction, and 67\% cost reduction while maintaining accuracy parity.

The centralized orchestration approach eliminates brittle conditional logic and cascading failures characterizing predetermined hierarchical systems, replacing them with autonomous decision-making that adapts to query complexity through adaptive routing strategies. The framework's modular design enables integration within larger systems or microservice deployment.

This work establishes foundations for scalable AI deployment prioritizing both technical excellence and economic sustainability---demonstrating that intelligent coordination of specialized components outperforms monolithic solutions across all measured dimensions.

\bibliographystyle{plain}
\bibliography{references}

\end{document}